\documentclass[letterpaper, 10 pt, journal, twoside]{IEEEtran}
\makeatletter
\let\NAT@parse\undefined
\makeatother
\usepackage[colorlinks]{hyperref}
\hypersetup{colorlinks=true,linkcolor=red,citecolor=blue,urlcolor=magenta}

\usepackage[utf8]{inputenc}
\usepackage[T1]{fontenc}
\usepackage{graphicx}
\usepackage{multirow}
\usepackage{amsmath}
\usepackage{amssymb} 
\usepackage{verbatim}
\usepackage{makecell}
\usepackage{caption}
\usepackage{pifont}
\usepackage{booktabs}
\usepackage[table,dvipsnames]{xcolor} % for rowcolor
\usepackage{amsmath} %for align
\usepackage{tabularx}
\usepackage{array}

\begin{document}
\bstctlcite{BSTcontrol}
\title{\LARGE\bf BLaFA: Bridging Language to 3D Functional Affordances for Dexterous Robotic Grasping in Tool Use}
\author{Fan Yang$^{1}$, Wenrui Chen$^{2,3}$, Guorun Yan$^{2}$, Ruize Liao$^{2}$, Wanjun Jia$^{2}$, Dongsheng Luo$^{2}$, Jianwen Hu$^{1}$,\\ Jiacheng Lin$^{4}$, Kailun Yang$^{2,3}$, Zhiyong Li$^{2,3}$, and Yaonan Wang$^{2,3}$%
\thanks{This work was partially supported by the Hunan Science Fund for Distinguished Young Scholars under Grant 2024JJ2027, the National Natural Science Foundation of China under Grants 62273137, 62473139, and No. U23A20341, and the Hunan Provincial Research and Development Project under Grant 2025QK3019.
\textit{(Corresponding author: Wenrui Chen.)}}
\thanks{$^{1}$The author is with the School of Artificial Intelligence, Changsha University of Science and Technology, Changsha 410118, China.}%
\thanks{$^{2}$The authors are with the School of Artificial Intelligence and Robotics, Hunan University, Changsha 410012, China. (E-mail: ysyf293@hnu.edu.cn, chenwenrui@hnu.edu.cn.)}%
\thanks{$^{3}$The authors are also with the National Engineering Research Center of Robot Visual Perception and Control Technology, Hunan University, Changsha 410082, China.}%
\thanks{$^{4}$The author is with the Key Laboratory of Advanced Manufacturing Technology of the Ministry of Education, Guizhou University, Guiyang 550025, China.}%
}

\maketitle
\begin{abstract}
In industrial assembly, equipment maintenance, and everyday tool-use scenarios, robots need not only to grasp objects stably but also to establish task-appropriate contact configurations according to task intentions and tool functionality. However, existing methods lack explicit mappings from open-vocabulary instructions to 3D functional contact structures and dexterous grasp configurations, limiting robots' adaptability to diverse tools and functional tasks. To address this limitation, we propose BLaFA (Bridging Language to 3D Functional Affordances for Dexterous Robotic Grasping), a structured framework for language-driven functional dexterous grasping. First, a Knowledge-guided Language Parsing (KLP) module leverages structured knowledge to transform natural-language instructions into semantic sextuples of manipulation constraints. Second, a 3D Gaussian Splatting based hierarchical semantic representation and a Triangular Functional Point Localization (TriLocation) module are developed to localize 3D functional contact keypoints under task–tool topology constraints. Finally, a 3D Keypoint Grasp Matrix Transformation (KGT3D+) module maps the localized keypoints and semantic constraints to executable wrist poses and finger-level joint and force commands. Evaluations on $100$ open-vocabulary language instructions, $14$ task–tool combinations covering four functional intents, and real-world robotic experiments in representative scenes demonstrate that BLaFA can effectively translate natural-language instructions into task-aligned 3D functional contact structures and corresponding dexterous grasps, thereby enabling language-driven robotic tool use across diverse tasks. The source code will be made publicly available. Code will be publicly available at \url{https://github.com/PopeyePxx/BLaFA}.
\end{abstract}

\section{Introduction}

Dexterous hands enable robots to interact with objects in human environments and manipulate tools designed for human use~\cite{zhong2026dexgraspvla, guo2025grasp, ma2026iep}. In industrial assembly, equipment maintenance, and everyday tool-use scenarios, such operations are often specified through natural-language instructions, requiring robots to understand task intentions and translate them into appropriate physical interactions. A fundamental capability underlying these interactions is functional dexterous grasping. Beyond simply achieving a stable grasp~\cite{wang2026rs2r}, it requires selecting appropriate contact regions and hand configurations according to the intended use of an object~\cite{zhu2023toward}.
For example, operating a drill requires positioning the finger responsible for functional operation at its trigger while maintaining a stable supporting grasp.

\begin{figure}[t!]
\centerline{\includegraphics[width=0.5\textwidth]{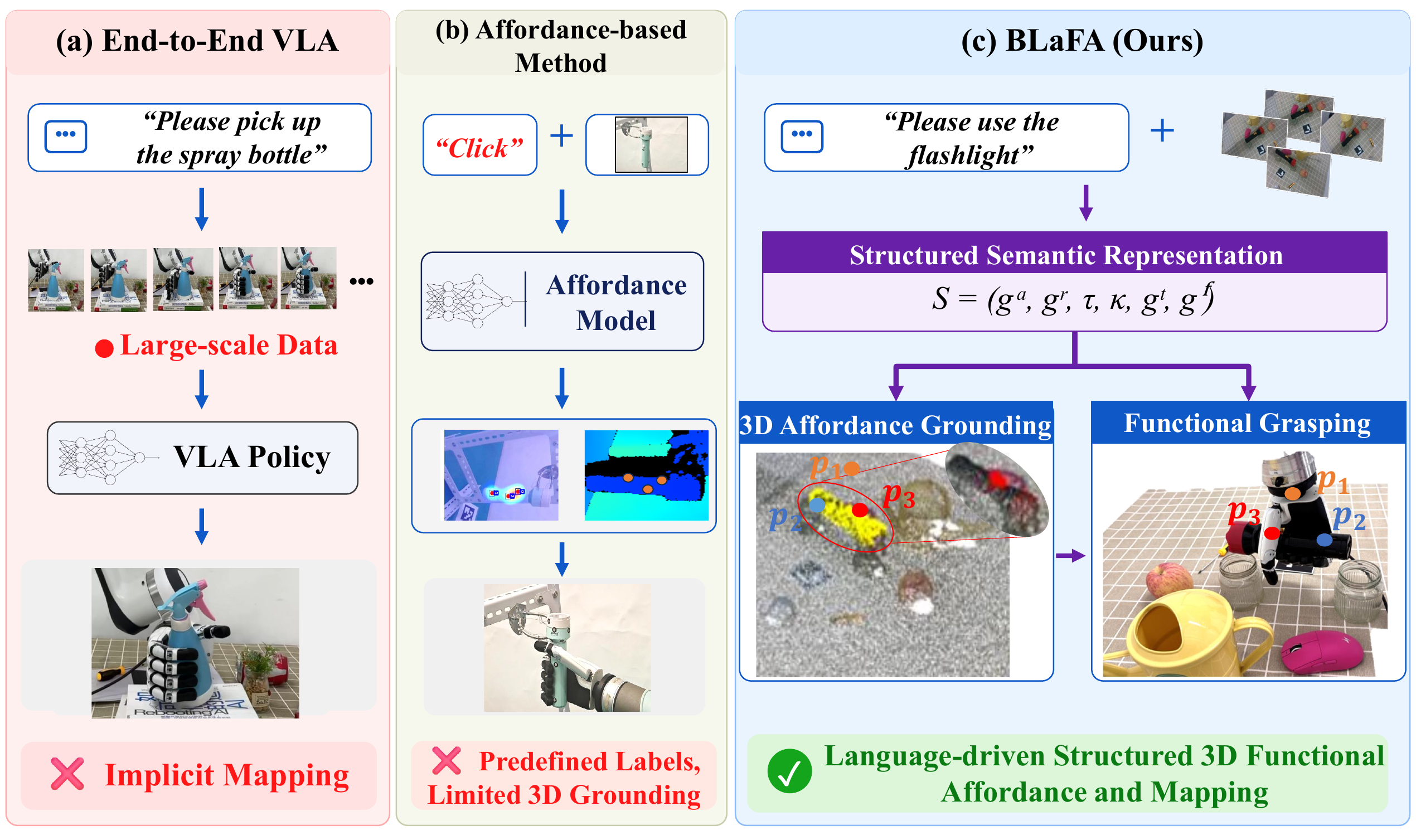}}
\captionsetup{font=small}
\caption{\small Comparison of representative paradigms: (a) end-to-end VLA methods without explicit representations of functional contact constraints; (b) affordance-based methods relying on predefined labels and 2.5D object-centric localization; and (c) BLaFA (ours), explicitly grounding language instructions into 3D functional contact structures for dexterous grasping.}
\label{intro}
\vskip-2ex
\end{figure}

Early studies on functional tool grasping primarily explored hand--object contact representations and dexterous hand configuration modeling~\cite{zhu2023toward, brahmbhatt2019contactdb, brahmbhatt2019contactgrasp, lakshmipathy2022contact,  DBLP:conf/iros/BrahmbhattHHF19}, laying the foundation for functional grasp synthesis.
With the development of Large Language Models (LLMs) and related foundation models, end-to-end Vision-Language-Action (VLA) models~\cite{brohan2023rt, liu2024robomamba, wen2025diffusionvla, he2025dexvlg} learn mappings from natural-language instructions and visual observations to robot actions, demonstrating promising performance on simple manipulation tasks, as shown in Fig.~\ref{intro}(a). However, they typically rely on large-scale training data, and their implicit, action-oriented representations make it difficult to explicitly capture the complex relationships between task semantics and dexterous grasping.

Another category of hierarchical methods~\cite{chen2022towards_human_level, huangrekep} decomposes manipulation into multiple modules to improve system controllability.
Among them, affordance-based methods~\cite{wei2025afforddexgrasp, li2025learningeccv, yang2025learning, yang2025multi} explicitly model the relationships among task semantics, object functional regions, and hand-object contacts, providing structured representations for functional grasp generation. As illustrated in Fig.~\ref{intro}(b), a recent study~\cite{yang2025multi} proposed a multi-keypoint affordance representation that establishes geometric correspondences between object functional regions and dexterous hand configurations. However, some existing methods rely on predefined affordance categories or 2.5D object-centric representations, making it difficult to directly map open-vocabulary instructions to spatially consistent 3D functional contact structures.
Furthermore, SparseGrasper~\cite{zhang2026sparse} and ToolBot~\cite{wei2023toolbot} have explored 3D grasp localization and keypoint-based tool-use representations, respectively, but their representations remain insufficient to capture the fine-grained contact relationships required for functional dexterous grasping. SayFuncGrasp~\cite{li2025language} establishes connections among language, perception, and functional grasping through a modular design, but relies on a learning-based solver to generate finger-level grasp configurations, making it difficult to explicitly characterize the correspondence between semantic constraints and the resulting grasp configurations.

Based on these observations, we investigate the following question: How can natural-language instructions be transformed into structured semantic constraints, grounded in hierarchical 3D functional representations, and further converted into geometrically consistent dexterous grasp configurations?

To address this challenge, we propose BLaFA (Bridging Language to 3D Functional Affordances for Dexterous Robotic Grasping), a structured framework for language-driven functional dexterous grasping, as illustrated in Fig.~\ref{intro}(c).
Specifically, BLaFA first transforms natural-language instructions into a structured semantic sextuple \(\mathcal{S}\), capturing task- and grasp-related constraints. These semantic constraints are then grounded in 3D to identify functional regions and localize task-relevant contact keypoints. Finally, the resulting 3D contact structure is integrated with grasp-type and force constraints to generate executable grasp configurations, including wrist poses and finger-level control commands.
Through this process, BLaFA establishes an explicit mapping from natural-language instructions to 3D functional affordances and task-aligned dexterous grasping.

First, we introduce a Knowledge-guided Language Parsing (KLP) module that integrates an LLM with structured domain knowledge and representative instruction examples to convert natural-language descriptions into explicit semantic constraints for subsequent 3D localization and grasp generation. Given an instruction, KLP generates a semantic sextuple: 
$\mathcal{S} = (g^a, g^r, g^t, g^f, \tau, \kappa)$, where $g^a$ denotes the functional grasp region, $g^r$ specifies the functional finger role, $g^t$ represents the grasp type, $g^f$ indicates the required interaction force, $\tau$ describes the tool topology, and $\kappa$ denotes the task intent. 

Next, we propose a Triangular Functional Point Localization (TriLocation) module to establish hierarchical semantic representations and localize task-oriented contact keypoints in 3D scenes. Based on multi-view RGB observations, we construct a 3D Gaussian Splatting representation enriched with object-level and part-level semantic features. To improve fine-grained functional part representation, we introduce a Hierarchical Semantic Extraction (HSE) strategy that separates object-level and part-level features while preserving contextual information during feature extraction.
Building upon this hierarchical semantic Gaussian field, TriLocation identifies the functional anchor point according to the grasp affordance $g^a$ and predicts the corresponding supporting keypoints according to the functional finger role $g^r$. Furthermore, by incorporating task intent $\kappa$ and tool topology $\tau$, we construct a task-conditioned local coordinate frame to constrain the spatial configuration of the three functional keypoints.

Finally, we develop a 3D Keypoint Grasp Matrix Transformation (KGT3D+) module to extend keypoint-based grasp transformation to 3D pose construction and finger-level control.
Given the three contact keypoints from TriLocation, KGT3D+ computes a task-consistent wrist pose through geometric transformations and maps the grasp type $g^t$ and force parameter $g^f$ to finger joint configurations and force profiles using a predefined functional grasp library.
The main contributions of this work are summarized as follows:
\begin{enumerate}
    \item An interpretable and modular 3D language-to-grasp framework, termed BLaDA, is proposed, where a structured semantic sextuple is employed to connect natural-language instructions, 3D functional perception, and dexterous grasp generation.

    \item A Knowledge-guided Language Parsing (KLP) mechanism is developed, in which a structured knowledge graph is incorporated to guide LLMs in transforming open-vocabulary instructions into structured dexterous grasp constraints.

    \item A 3DGS-based hierarchical functional semantic representation and a Triangular Functional Point Localization (TriLocation) module are proposed, where task--tool topology constraints are incorporated to establish the mapping from 2D semantic regions to 3D functional contact structures.
\end{enumerate}

\section{Related Work}

\subsection{Affordance Grounding in 3D}

Functional affordance modeling is essential for task-oriented manipulation, requiring accurate identification and localization of semantically meaningful regions in 3D space. Traditional RGB-, RGB-D-, and point cloud-based methods~\cite{luo2022learning, li2023locate, zhu2025grounding} often suffer from resolution limitations, data sparsity, or reliance on predefined affordance categories, restricting their applicability to fine-grained manipulation.
Although neural implicit representations~\cite{shen2023distilled} improve surface continuity, limitations in real-time interaction and interpretability remain.
In contrast, 3D Gaussian Splatting (3DGS)~\cite{kerbl20233d} provides an efficient and differentiable scene representation with high-fidelity geometry and rich appearance information. Recent works, including GaussianGrasper~\cite{zheng2024gaussiangrasper}, GraspSplats~\cite{jigraspsplats}, and SparseGrasper~\cite{zhang2026sparse}, have demonstrated the potential of 3DGS for semantic grounding and language-guided robotic grasping. However, these methods primarily focus on conventional grasp generation, leaving the fine-grained functional contact relationships and finger-level configurations required for dexterous grasping insufficiently addressed.

In this work, we extend 3DGS-based scene representations to support natural-language-guided functional affordance grounding, enabling 3D localization of task-relevant contact points and executable dexterous grasp generation.

\subsection{Language-Guided Robotic Manipulation}
Mapping natural language instructions to executable robotic actions has attracted increasing attention.
Existing approaches mainly follow two paradigms: end-to-end Vision-Language-Action (VLA) learning~\cite{brohan2023rt, liu2024robomamba, wen2025diffusionvla, he2025dexvlg, ahn2022can}, which jointly maps language and visual observations to robot actions, and hierarchical approaches~\cite{ chen2022towards_human_level, huangrekep}, which separate high-level semantic reasoning from low-level action execution. Despite their progress, explicitly establishing fine-grained correspondences between task semantics, 3D functional contact regions, and finger-level grasp configurations remains challenging.

\begin{figure*}[!t]
\centerline{\includegraphics[width=0.9\textwidth]{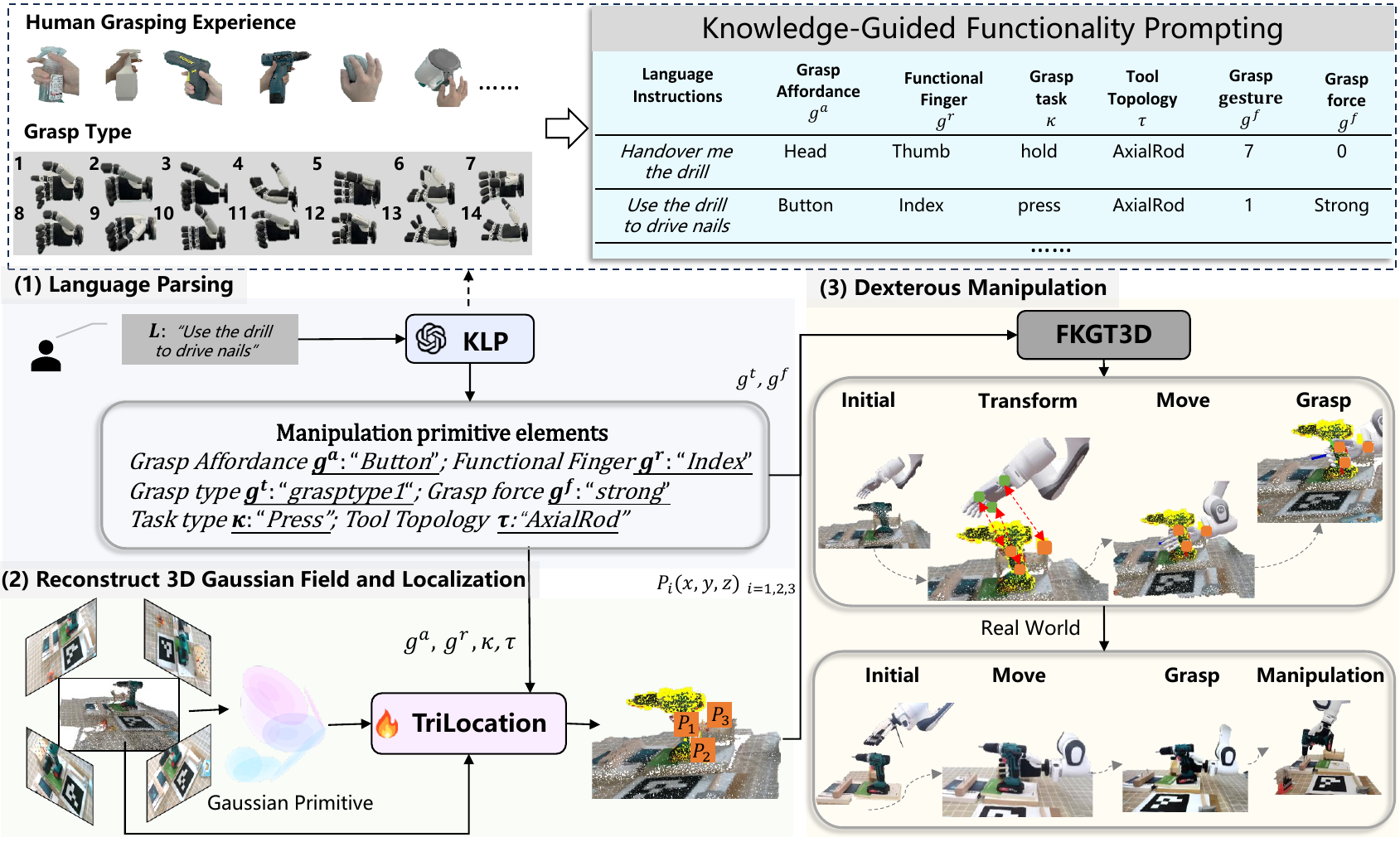}}
\captionsetup{font=small}
\caption{\small Overview of BLaFA. 
The top illustrates knowledge-guided prompt construction with two example instructions. The bottom presents the three-stage pipeline: (1) Language parsing (blue); (2) 3D reconstruction and localization (green); (3) Dexterous grasping (yellow).}
\label{pipline}
\vskip-2ex
\end{figure*}

Among existing methods, SayFuncGrasp~\cite{li2025language} is closely related to our work. It employs LLMs to infer functional grasp requirements but relies on a learned policy planner for finger-level control, leaving the mapping between semantic constraints and grasp configurations insufficiently explicit. 
In contrast, we introduce a structured semantic sextuple $\mathcal{S}=(g^a,g^r,g^t,g^f,\tau,\kappa)$ to encode language intent into explicit perceptual and control constraints, guiding subsequent 3D functional localization and dexterous grasp generation.

\subsection{Object Representation for Dexterous Grasping}

Traditional 6-DoF grasp representations~\cite{yu2023robotic, srivastava2014combined, tyree20226, wen2024foundationpose} are insufficient to describe the complex contacts required for dexterous grasping.
Previous studies have explored functional contact representations~\cite{brahmbhatt2019contactdb, cao2022efficient, zhu2023toward}, but accurate perception and generalization remain challenging. Generating executable grasps from localized semantic regions further requires accurate estimation of grasp orientations and geometric constraints.
Existing approaches explore discretized orientation search~\cite{chu2018real} and geometric analysis~\cite{lundell2021ddgc}, such as PCA-based pose alignment using point clouds~\cite{choi2018learning}. Keypoint-based methods, such as ToolBot~\cite{wei2023toolbot}, represent task-oriented grasping and manipulation through oriented keypoints.
Meanwhile, DexFuncGrasp~\cite{hang2024dexfuncgrasp} and MKA~\cite{yang2025multi} model wrist-object relationships using three-point geometric constraints. However, these methods often rely on depth-based geometric reconstruction, which can limit accurate localization in complex scenes. In contrast, we propose a novel paradigm based on 3D Gaussian Splatting (3DGS), which embeds grasp constraints directly into a dense geometric field, ensuring more robust functional localization and grasp synthesis.

\section{Methodology}
\textbf{Problem Formulation}
This study aims to achieve functional dexterous grasping in a 3D reconstructed scene using natural language instructions and multi-view observations.
Specifically, the goal is to infer: (i) a set of three keypoint coordinates $P = \{p_1, p_2, p_3\}$ in 3D space on the target object, and (ii) a grasp configuration $G = (R, T, J, F)$ that defines the wrist pose, joint configuration, and finger-level force distribution.

Formally, the proposed framework $\mathcal{M}$ takes as input a language instruction $L$ and a collection of RGB-D images $\Omega$, and outputs the grasp configuration $G$ and keypoint set $P$ as:
\begin{equation}
G, P = \mathcal{M}(L, \Omega).
\end{equation}

\textbf{Pipeline}
As shown in Fig.~\ref{pipline}, our framework consists of three stages: (1) Language parsing (blue), where KLP transforms the input instruction $L$ into structured grasp constraints $\mathcal{S}=(g^a,g^r,g^t,g^f,\tau,\kappa)$; (2) 3D reconstruction and functional localization (green), where TriLocation constructs a 3D Gaussian field with object--part hierarchical semantic information and localizes three functional keypoints conditioned on $(g^a,g^r,\tau,\kappa)$; and (3) Dexterous grasp generation (yellow), where KGT3D+ transforms the localized keypoints into hand--object grasp poses and generates finger-level configurations conditioned on $(g^t,g^f)$. Details are provided in Secs.~\ref{klp}, \ref{3dgs}, and \ref{sec:kgt}, respectively.

\begin{figure*}[t]
\centerline{\includegraphics[width=0.9\textwidth]{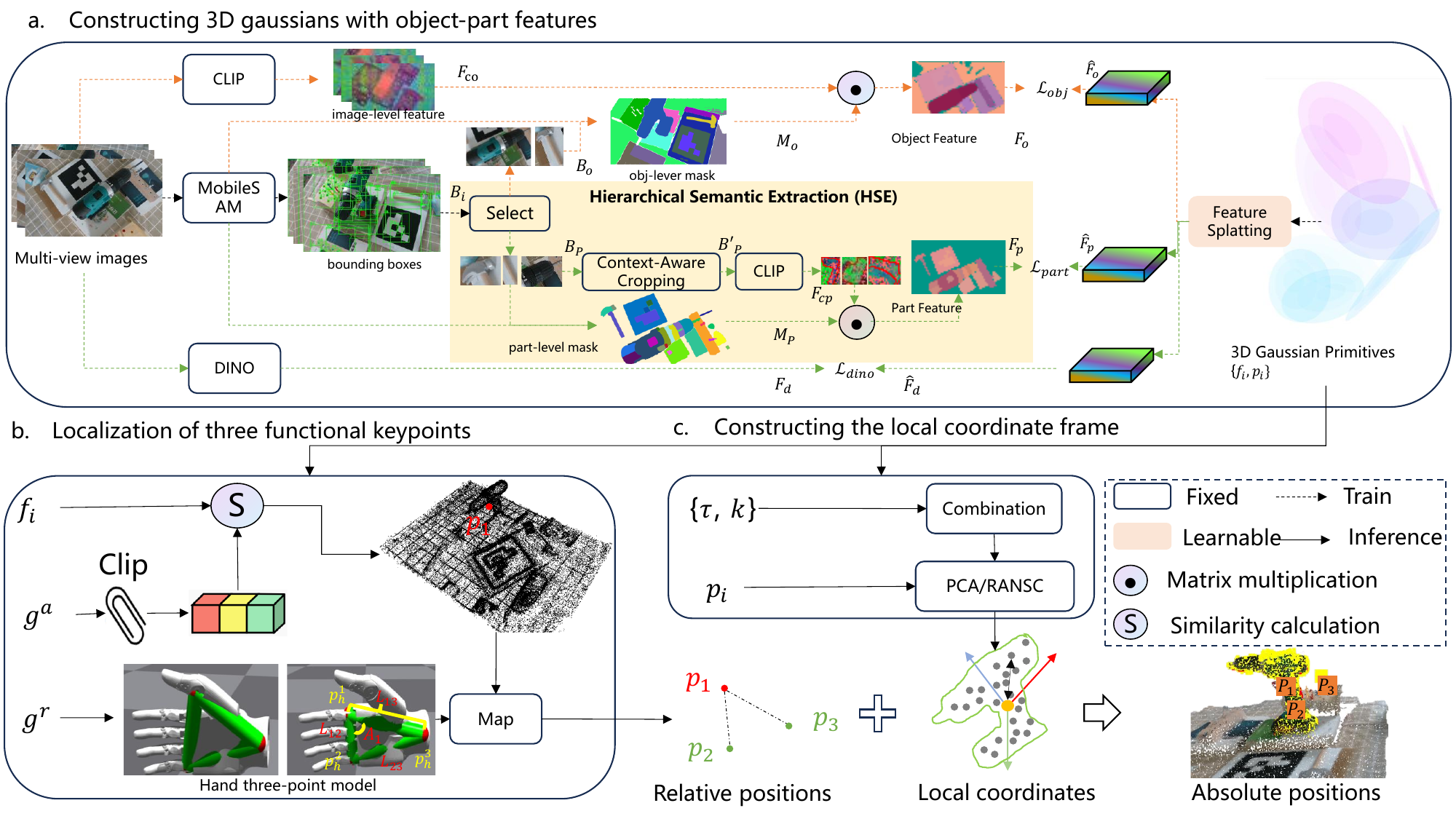}}
\captionsetup{font=small}
\caption{\small Overview of TriLocation. (a) HSE (yellow) separates object- and part-level regions through Select and Context-Aware Cropping, mitigating semantic suppression and drift. (b) CLIP similarity between $g^a$ and Gaussian features $f_i$ localizes the semantic anchor $p_1$. (c) An MLP predicts offsets $\Delta p_2$ and $\Delta p_3$ to form the grasp triangle $\{p_1,p_2,p_3\}$, supervised by edge-length and angle constraints.
}
\label{tri}
\vskip-2ex
\end{figure*}

\subsection{Knowledge-guided Language Parsing (KLP)}\label{klp}
We formulate functional grasping knowledge as a sextuple $\mathcal{S}=(g^a,g^r,g^t,g^f,\tau,\kappa)$, comprising four core primitives: grasp affordance region $g^a$, functional finger role $g^r$, grasp type $g^t$, and force level $g^f$, wich is following the tool-use-oriented F2F knowledge graph~\cite{yang2024task}. 

\begin{figure}[!t]
\centerline{\includegraphics[width=0.5\textwidth]{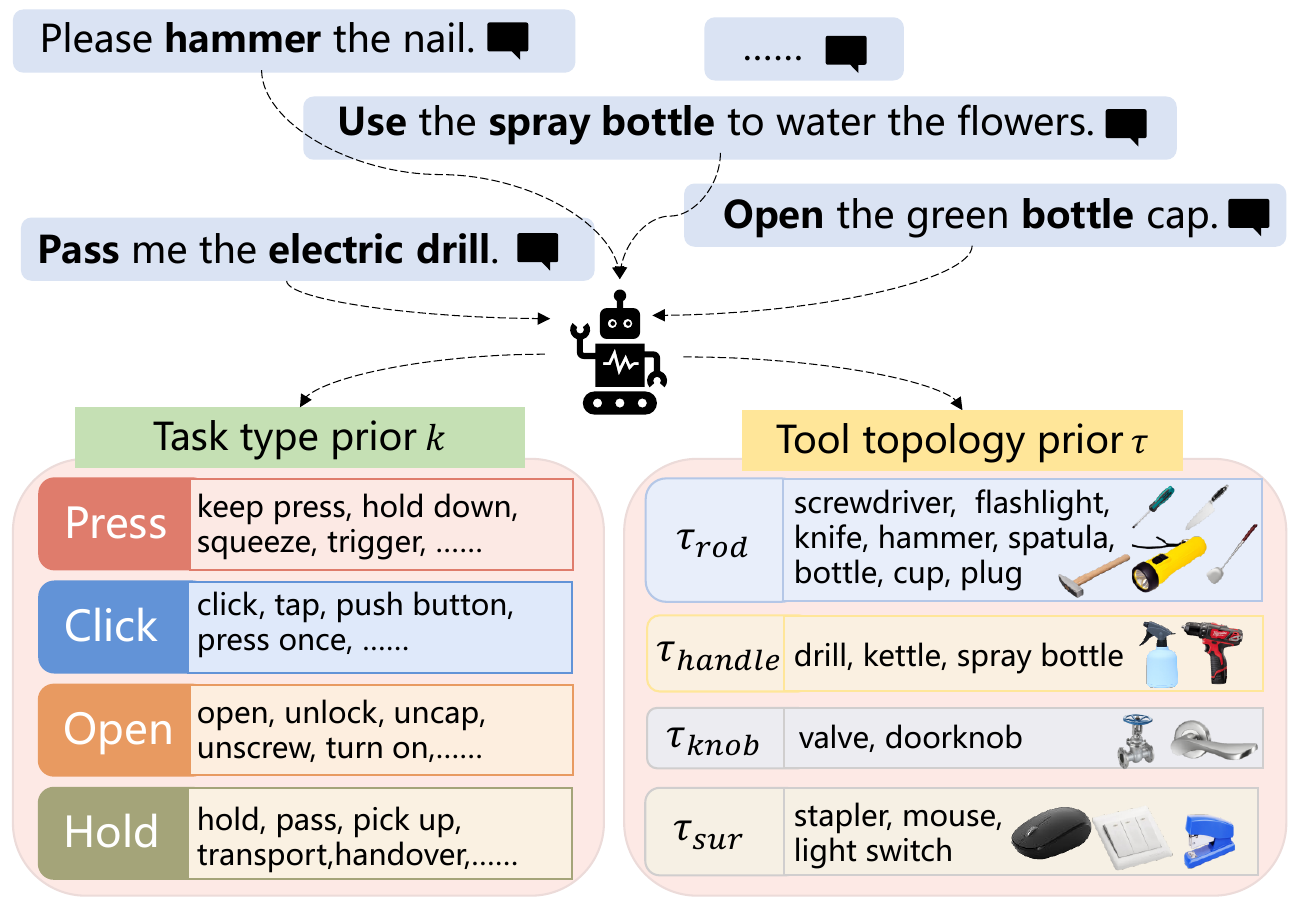}}
\captionsetup{font=small}
\caption{\small Extraction of verbs, intent phrases, and tool types from natural-language instructions $L$, followed by their classification into task intent and tool topology priors.}
\label{tasktool}
\vskip-3ex
\end{figure}

Specifically, to address ambiguity and control instability in 3D functional interactions, we further introduce tool topology prior $\tau$ and task intent prior $\kappa$, as illustrated in Fig.~\ref{tasktool}. Specifically, $\tau$ incorporates geometric features and human operating habits to classify objects into four categories: axial rods $\tau_{\mathrm{rod}}$, lateral handles $\tau_{\mathrm{handle}}$, knobs or wheels $\tau_{\mathrm{knob}}$, and slabs or surfaces $\tau_{\mathrm{surface}}$. Meanwhile, $\kappa$ maps instruction verbs to four atomic task types: $\mathrm{press}$, $\mathrm{click}$, $\mathrm{open}$, and $\mathrm{hold}$. The extracted tool types and task-related phrases are mapped to $\tau$ and $\kappa$, respectively, providing structured constraints for subsequent 3D localization and grasp generation.

Then we design a prompt with three key components: (i) role specification, defining the agent as an embodied intelligence; (ii) structured injection, providing the grasp taxonomy, task taxonomy, and tool topology models from the F2F knowledge base in a formalized manner; and (iii) in-context examples, using representative ``instruction--reasoning--tuple'' exemplars to guide the model in generating standardized outputs.

Finally, given an instruction $L$ and a language prompt $\Pi$, KLP maps them to a structured sextuple:
\begin{equation}
\mathcal{S}=(g^a,g^r,g^t,g^f,\tau,\kappa)=\mathrm{KLP}(L,\Pi).
\end{equation}
In implementation, KLP is driven by a large language model.

\subsection{Triangular Functional Point Localization Module (TriLocation)}\label{3dgs}

Previous studies~\cite{yang2025multi, hang2024dexfuncgrasp} have shown that three contact keypoints on an object, corresponding to the functional finger, the little finger, and the heel of the palm, can effectively constrain hand--object contact relationships. Accordingly, this module aims to achieve semantic-guided 3D localization of these three points using 3DGS. To this end, we propose the TriLocation module (Fig.~\ref{tri}), which consists of three steps: constructing a 3D Gaussian representation with object--part features, localizing three functional keypoints, and constructing a local coordinate frame.

\subsubsection{Constructing 3D Gaussians with Object--Part Features}

The main challenge in constructing 3D Gaussians with object–part features lies in addressing semantic suppression and semantic drift. To this end, we propose a Hierarchical Semantic Extraction (HSE) strategy, consisting of a Select Module and a Context-aware Cropping Module, to extract object-level and part-level semantic features.

As shown in Fig. 3(a), given an input image \(I\), we detect candidate bounding boxes \(\{B_k\}_{k=1}^N\) using YOLO~\cite{redmon2016you} and generate corresponding masks \(\{M_k\}_{k=1}^N\) using SAM~\cite{kirillov2023segment}. The Select Module partitions these masks into object-level regions \(\mathcal{M}_o\) and part-level regions \(\mathcal{M}_p\) based on the mask-to-box area ratio and a threshold \(\alpha\), distinguishing large object/background regions from localized functional parts.

For object-level feature extraction, we encode the full image using CLIP~\cite{radford2021learning} to obtain a dense feature map \(F_{\mathrm{co}}\), from which object-level features \(F_o\) are extracted. For part-level features, we adopt context-aware cropping to preserve global semantic information. Specifically, each part bounding box \(B_p=[x_1,y_1,x_2,y_2]\) is expanded by a padding ratio \(\gamma\):
\begin{equation}
B'_p=[x_1-\gamma w,\ y_1-\gamma h,\ x_2+\gamma w,\ y_2+\gamma h],
\end{equation}
where \(w\) and \(h\) denote the width and height of \(B_p\), respectively. The expanded crop is encoded by CLIP, and its central features are projected onto the corresponding SAM mask \(M_p\) to obtain part-level features \(F_p\). Meanwhile, DINO-v2 extracts image features \(F_d\) as additional supervision.

On the 3DGS side, each Gaussian primitive is assigned a low-dimensional latent feature \(\mathbf{f}_i\). These features are rendered through volumetric rendering to obtain \(\hat{\mathbf{F}}\), which is subsequently decoded by a shallow MLP into three branches: object-level features \(\hat{F}_o\), part-level features \(\hat{F}_p\), and DINO-v2 features \(\hat{F}_d\). The three branches are supervised by their corresponding image-derived features using cosine similarity losses, enabling hierarchical semantic distillation from multi-view images into the 3D Gaussian field. The overall objective is:
\begin{equation}
L=L_{\mathrm{obj}}+L_{\mathrm{part}}+\lambda L_{\mathrm{dino}},
\end{equation}
where \(\lambda\) balances the DINO-v2 supervision.

\begin{table}[t]
    \centering
    \caption{Task-Tool Topology Constraints for Local Coordinate Frames.}
    \label{tab:task_tool_topology_en}
    \footnotesize
    \renewcommand{\arraystretch}{1.08}
    \begin{tabular*}{\columnwidth}{@{\extracolsep{\fill}}ccll@{}}
        \toprule
        \textbf{Intent} & \textbf{Topology} &
        $\hat{z}$ (Primary Axis) &
        $\hat{y}$ (Hand Orientation) \\
        \midrule
        \multirow{4}{*}{Hold}
        & $\tau_{\mathrm{rod}}$ & Gravity dir. & Structural axis \\
        & $\tau_{\mathrm{handle}}$ & Radial dir. & Handle axis \\
        & $\tau_{\mathrm{knob}}$ & Surface normal & Tangential dir. \\
        & $\tau_{\mathrm{surface}}$ & Surface normal & Major in-plane axis \\
        \midrule
        Press & $\tau_{\mathrm{handle}}$ & Tool forward axis & Handle axis \\
        \midrule
        Open & $\tau_{\mathrm{knob}}$ & Rotation axis & Tangential dir. \\
        \midrule
        \multirow{2}{*}{Click}
        & $\tau_{\mathrm{surface}}$ & Click normal & Major in-plane axis \\
        & $\tau_{\mathrm{rod}}$ & Click normal & Major in-plane axis \\
        \bottomrule
    \end{tabular*}
\end{table}

\subsubsection{Localization of Three Functional Keypoints}

Given the hierarchical semantic 3D Gaussian field, TriLocation localizes three functional keypoints based on the KLP outputs, as shown in Fig.~\ref{tri}(b). First, the grasp-region semantics \(g^a\) is encoded by CLIP~\cite{radford2021learning} into a query vector, whose similarity \(S\) with each Gaussian feature \(f_i\) is calculated. Gaussians with similarity above a threshold \(\delta\) are clustered in 3D space, and the centroid of the highest-confidence cluster is selected as the semantic anchor \(p_1\).

Next, the grasp-relation semantics \(g^r\) guides the Hand three-point model to select a functional hand template \(\{p_{h1},p_{h2},p_{h3}\}\), corresponding to the functional fingertip, little finger, and wrist, respectively. The Map module then transfers the template to the object using the task-topology local coordinate frame (Sec.~\ref{tt}), aligning \(p_{h1}\) with the semantic anchor \(p_1\) and determining the remaining points \(p_2\) and \(p_3\) according to the template geometry. Their world coordinates are computed as:
\begin{equation}
\mathbf{p}_2 = \mathbf{p}_1 + R
\begin{bmatrix}
0\\
-L_{12}\sin A_1\\
-L_{12}\cos A_1
\end{bmatrix},
\quad
\mathbf{p}_3 = \mathbf{p}_1 + R
\begin{bmatrix}
0\\
0\\
-L_{13}
\end{bmatrix},
\end{equation}
where \(R\in\mathbb{R}^{3\times3}\) is determined by the task-topology rules, with its columns representing the local axes \((\hat{x},\hat{y},\hat{z})\) in the world coordinate frame.

\subsubsection{Constructing the local coordinate frame}\label{tt}

To resolve pose ambiguity and avoid mirrored contact configurations, we construct a task-conditioned local coordinate frame by coupling task-intent prior \(\kappa\) with tool-topology prior \(\tau\), as illustrated in Fig.~\ref{tri}(c). As summarized in Table~\ref{tab:task_tool_topology_en}, each task–tool combination specifies a primary task axis \(\hat{z}\) and a hand orientation axis \(\hat{y}\). These axes are estimated from point clouds using topology-specific geometric fitting: RANSAC~\cite{raguram2012usac} for rod, handle, and knob axes, and PCA~\cite{mackiewicz1993principal} for surface normals and dominant in-plane directions of planar structures. Task-dependent geometric cues, such as radial and tool-forward directions, are further incorporated to determine grasp depth and orientation.

Finally, \(\hat{y}\) is projected onto the plane orthogonal to \(\hat{z}\), and the remaining axis is obtained by \(\hat{x}=\hat{y}\times\hat{z}\), yielding a consistent right-handed coordinate frame for defining the three functional contact keypoints.

\subsection{3D Keypoint-based Grasp Matrix Transformation (KGT3D+)}\label{sec:kgt}

KGT3D+ integrates geometric contact constraints and functional grasp semantics to generate executable dexterous grasp configurations. Given the three 3D contact keypoints $\{p_1,p_2,p_3\}$ from TriLocation, we construct object and hand local coordinate frames based on the corresponding point triplets. The rigid transformation aligning these two frames yields the 3D grasp pose $G=(R,T)$, following the formulation in MKA~\cite{yang2025multi}.

Meanwhile, to incorporate task-specific grasp requirements, the grasp type $g^t$ and force label $g^f$ parsed by KLP are mapped to the finger joint configuration $J$ and force profile $F$ through the functional grasp library F2F~\cite{yang2024task}. Finally, the geometric pose and functional control parameters are integrated into a unified grasp configuration $\mathcal{C}=(R,T,J,F)$, enabling functional dexterous grasp execution.

\section{Experiments}

In this section, we evaluate the proposed framework from three aspects. First, we assess the KLP module's ability to extract structured manipulation constraints from natural language. Second, we evaluate the TriLocation module in terms of 2D semantic localization and 3D functional keypoint localization. Finally, real-world robot experiments are conducted to evaluate the integrated framework, where KGT3D+ maps localized contact keypoints into grasp poses and finger-level configurations with joint and force parameters.

\subsection{Setup}

\subsubsection{Scenes, Objects, and Devices} 
\begin{figure}[!t]
    \centering
\includegraphics[width=1\linewidth]{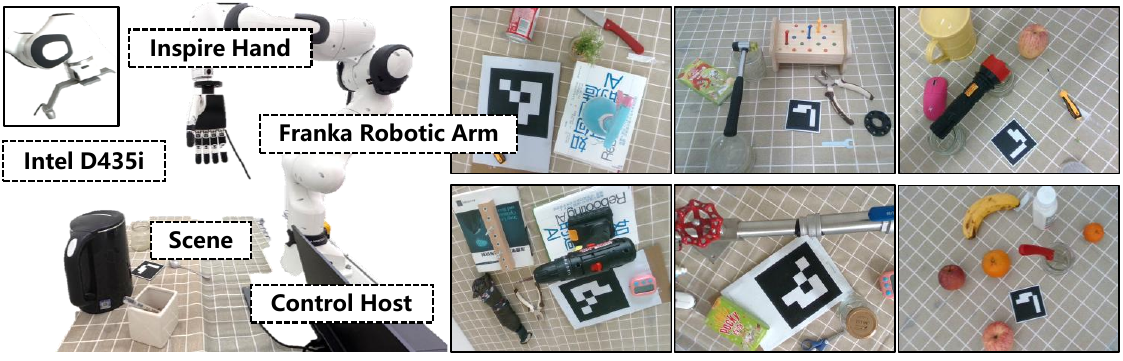}
    \caption{Real-world experiment setting and 6 typical scenarios demonstration.}  \label{fig:hardware}
\vskip-2ex
\end{figure}

Our experimental setup is illustrated on the left side of Figure~\ref{fig:hardware}. It comprises a Franka Emika robot arm equipped with a 6-DOF (degree-of-freedom) dexterous Inspire Hand for executing grasping tasks, and an Intel RealSense D435i camera for multi-view image acquisition. 
We configured over 10 open tabletop scenarios using $18$ types of tools from the representative dexterous grasping datasets FAH~\cite{ yang2024task, yang2025learning}. Six typical scenarios are showcased on the right of Fig.~\ref{fig:hardware}, based on which a total of $100$ language-guided manipulation trials were conducted. An NVIDIA RTX 3090 GPU was utilized for 3D reconstruction and the training of the TriLocation module.

\begin{figure*}[ht]
    \centering
\includegraphics[width=0.9\linewidth]{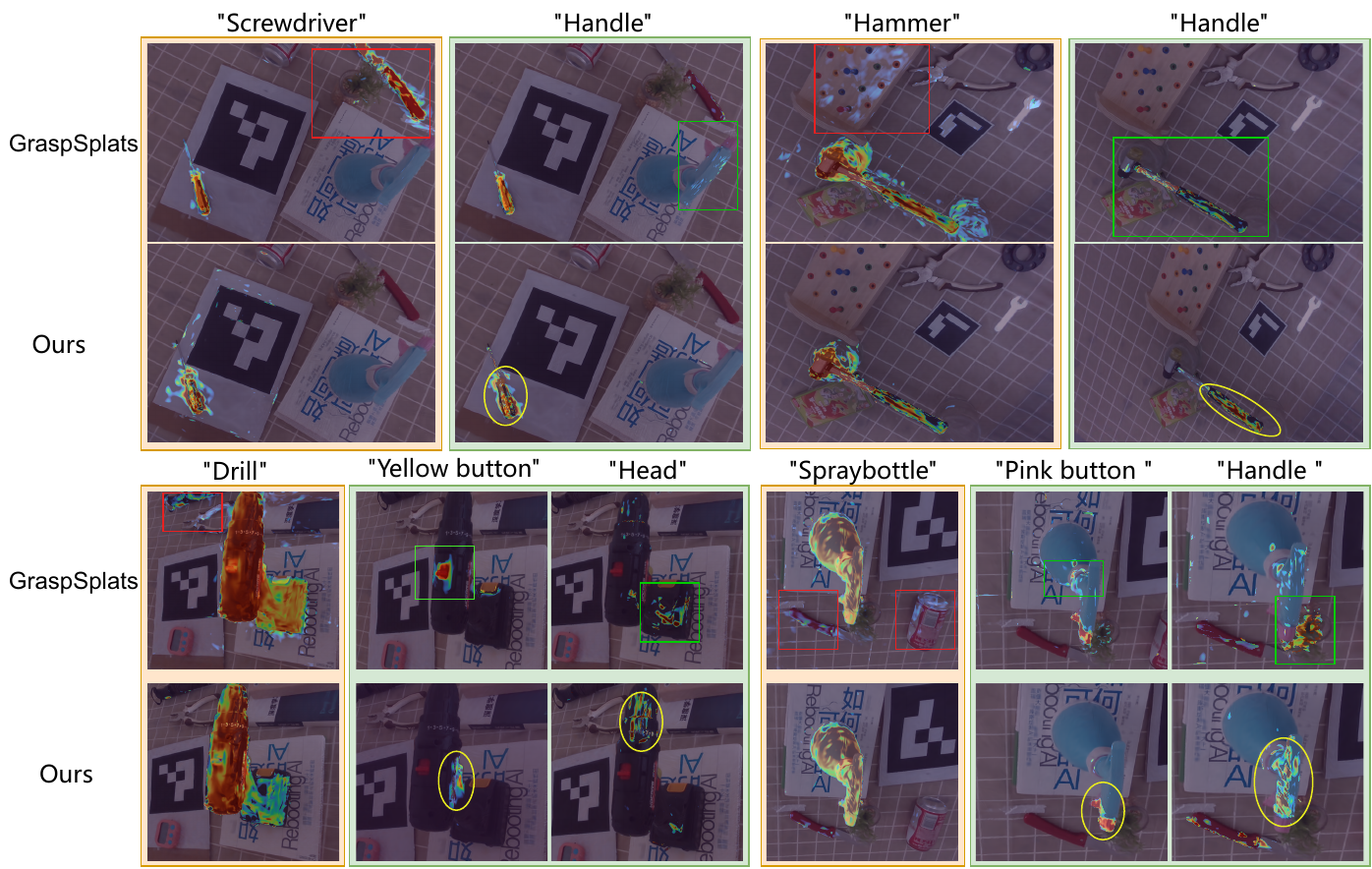}
    \caption{
    Language-guided relevance maps projected from 3D Gaussian features onto 2D images. Orange and green panels denote object- and part-level relevance, respectively. Red/green boxes highlight erroneous or ambiguous responses of GraspSplats~\cite{jigraspsplats}, while yellow ellipses indicate more compact and complete part-level responses of BLaFA.
}
    \label{fig:heatmap}
\end{figure*}

\subsubsection{Evaluation Metrics} 
For consistent mathematical notation, let $H$ and $H*$ denote the predicted and ground-truth affordance heatmaps, respectively, and $M$ be the binary mask of the target part. Let $P_i$, $G_i$, and $M_i$ represent the $i$-th pixel values, with $N$ being the total number of pixels. The specific metrics are defined as follows:

\begin{itemize}
    \item Language Reasoning Accuracy (LRA): 
    Evaluates the correctness of the system in inferring fundamental manipulation elements from natural language instructions. It is defined as the ratio of correctly inferred cases to the total number of test cases.
    
    \item 2D Part-level Localization Metrics: 
    To quantitatively evaluate the precision of 2D part-level feature extraction used to constrain 3D rendering, we adopt the following metrics:
    \begin{itemize}
        \item Mean Absolute Error (MAE): Measures the pixel-wise average deviation between $P$ and $G$: $MAE = \frac{1}{N} \sum_{i=1}^{N} |H_i - H*_i|$.
        \item Precision Energy ($P_{En}$): Quantifies the concentration of predicted energy within the target part mask $M$: $P_{En} = \frac{\sum H_i M_i}{\sum H_i}$.
        \item Affordance Grounding Metrics: Following \cite{luo2022learning, li2023locate}, we introduce KL Divergence (KLD), Similarity (SIM), and Normalized Scanpath Saliency (NSS) to assess the distributional alignment between the predicted heatmaps and the ground-truth part regions.
    \end{itemize}
    
    \item Localization Success Rate (LSR): 
  Whether the 3D coordinate deviations between keypoints are within predefined thresholds.
    
    \item Functional Grasp Success Rate (FSR):
    It measures the proportion of successful functional grasping tasks completed during physical trials.
\end{itemize}
\subsection{KLP-Based Language Parsing Evaluation}
To assess the contribution of structured domain knowledge, we evaluate three representative large language models: ChatGPT4.0~\cite{achiam2023gpt}, DeepSeekv3~\cite{liu2024deepseek}, and Gemini2.5~\cite{comanici2025gemini}, with and without the proposed KLP module. The evaluation covers six manipulation elements ($g^a, g^t, g^f, g^r, \tau, \kappa$), using Language Reasoning Accuracy (LRA) as the metric. As shown in Table~\ref{table_klp}, three observations can be made:

\textbf{Consistent improvement in average accuracy.} Integrating KLP improves the average LRA of all three models, from $0.508$ to $0.753$ for ChatGPT4.0~\cite{achiam2023gpt}, from $0.540$ to $0.743$ for DeepSeekv3~\cite{liu2024deepseek}, and from $0.542$ to $0.745$ for Gemini2.5~\cite{comanici2025gemini}, corresponding to an average improvement of approximately $21.7$ percentage points.

\textbf{Larger gains in grasp type and force prediction.} The most substantial improvements are observed in $g^t$ and $g^f$, which represent grasp type and force level, respectively. For example, ChatGPT4.0~\cite{achiam2023gpt} improves its $g^t$ accuracy from $0.13$ to $0.81$, while Gemini2.5~\cite{comanici2025gemini} achieves the highest $g^r$ accuracy of $0.80$. These results highlight the benefits of structured knowledge in domain-specific manipulation reasoning.

\textbf{Best overall performance with ChatGPT4.0.} Among the evaluated models, ChatGPT4.0 with KLP achieves the highest average LRA of $0.753$, although the best performance on individual elements varies across models.

Overall, the results demonstrate that KLP consistently improves average structured parsing accuracy across the three evaluated LLM backbones.

\begin{table}[!t]
    \centering
    \setlength{\tabcolsep}{2.5pt} % reduce column spacing
    \renewcommand{\arraystretch}{1.05} % adjust row height
    \caption{Comparison of language reasoning accuracy (LRA) across three LLMs with (w/) and without (w/o) the KLP module.
    Best results are highlighted in bold.}
    \label{table_klp}
    \begin{tabular}{lcccccc|c}
        \toprule
        \textbf{Model} & \(g^a\) & \(g^t\) & \(g^f\) & \(g^r\) & \(t\) & \(k\) & \textbf{Avg.} \\
        \midrule
        ChatGPT4.0~\cite{achiam2023gpt} (w/) & 0.65 & 0.81 & \textbf{0.74} & 0.74 & 0.72 & \textbf{0.86} & \textbf{0.753} \\
        ChatGPT4.0~\cite{achiam2023gpt} (w/o) & 0.43 & 0.13 & 0.45 & 0.51 & 0.72 & 0.81 & 0.508 \\
        \midrule
        DeepSeekv3~\cite{liu2024deepseek} (w/) & 0.62 & 0.80 & \textbf{0.74} & 0.79 & 0.73 & 0.78 & 0.743 \\
        DeepSeekv3~\cite{liu2024deepseek} (w/o) & 0.42 & 0.18 & 0.46 & 0.62 & \textbf{0.77} & 0.79 & 0.540 \\
        \midrule
        Gemini2.5~\cite{comanici2025gemini} (w/) & \textbf{0.70} & \textbf{0.85} & 0.64 & \textbf{0.80} & 0.66 & 0.82 & 0.745 \\
        Gemini2.5~\cite{comanici2025gemini} (w/o) & 0.51 & 0.22 & 0.50 & 0.65 & 0.59 & 0.78 & 0.542 \\
        \bottomrule
    \end{tabular}
\end{table}

\subsection{Performance Evaluation of TriLocation}

We evaluate the proposed TriLocation module from both qualitative visualization and quantitative analysis, and verify its effectiveness and advantages in two aspects: part-level feature extraction and three-dimensional localization of functional keypoints.

\textbf{Qualitative Comparison.}
We first compare our method with GraspSplats~\cite{jigraspsplats} in object- and part-level semantic rendering to evaluate the proposed HSE module. As shown in Fig.~\ref{fig:heatmap}, GraspSplats suffers from semantic cross-talk at the object level (red boxes), activating irrelevant but semantically similar regions, and part-level semantic diffusion (green boxes), where local queries often spread across the entire object. In contrast, our method produces clearer object boundaries and more compact and complete part-level responses. This improvement is attributed to HSE, which explicitly decouples object- and part-level features, reducing the interference of global object semantics on local part representations and providing more reliable semantic priors for functional region localization.

\begin{figure}[t]
    \centering
\includegraphics[width=1\linewidth]{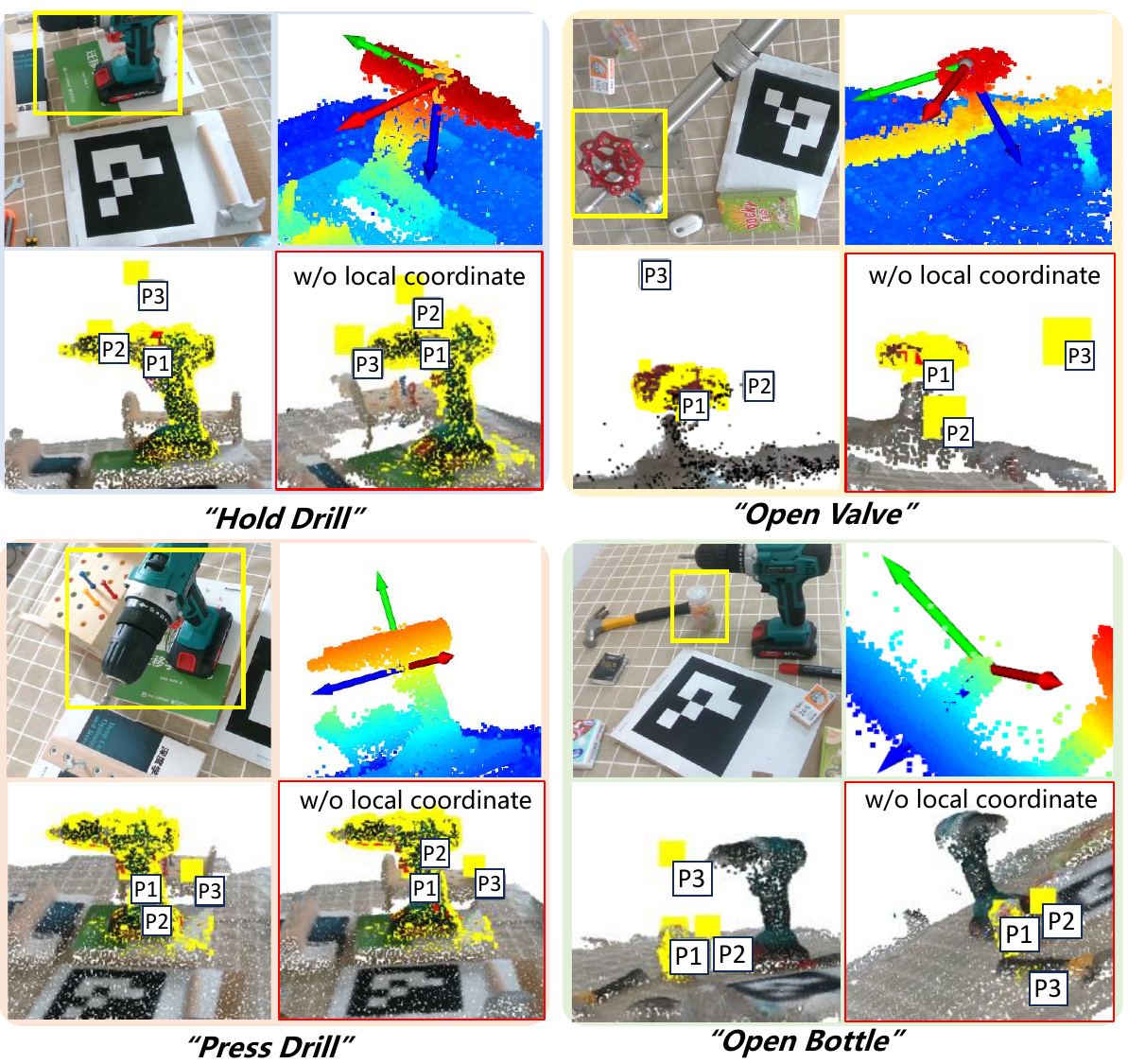}
    \caption{
    Visualization of the effect of the local coordinate system on 3D functional keypoint localization. Each example presents, in order: the input image, a 3D visualization of the local coordinate system (red/green/blue indicate the $x/y/z$ axes), the predicted three-point structure under the local-coordinate constraint, and the results without this constraint (red boxes).}
    \label{fig:3d_visualization}
\end{figure}

We further evaluate the effect of the local coordinate frame on 3D functional keypoint localization. As shown in Fig.~\ref{fig:3d_visualization}, we compare four task-tool combinations with and without local coordinate constraints. Without these constraints, the predicted triangular structures often exhibit geometric disorder and semantic misalignment. 
For example, in the ``Hold Drill'' and ``Open Bottle'' tasks, the inaccurate wrist point $p_3$ leads to incorrect approach directions. 
In contrast, our task-conditioned coordinate frame establishes consistent geometric references based on 3D scene structures, improving the geometric and semantic consistency of the predicted contact configurations.

\begin{table}[!t]
\centering
\caption{Quantitative comparison of 2D Part Localization Performance. Our method shows significant improvements in localization accuracy and energy focus across different object scales.}
\label{tab:performance_comparison}
\renewcommand{\arraystretch}{1.3} % 
\resizebox{\columnwidth}{!}{%
\begin{tabular}{llccccc}
\toprule % 
\textbf{Query Item} & \textbf{Model} & \textbf{MAE} $\downarrow$ & \textbf{P\_{En}} $\uparrow$ & \textbf{KLD} $\downarrow$ & \textbf{SIM} $\uparrow$ & \textbf{NSS} $\uparrow$ \\ \midrule
\multirow{2}{*}{Hammer Handle} & GraspSplats~\cite{jigraspsplats} & 0.0141 & 0.2798 & 14.1642 & 0.1449 & 1.2916 \\
 & \textbf{Ours} & \textbf{0.0127} & \textbf{0.5962} & \textbf{13.0819} & \textbf{0.2040} & \textbf{2.7209} \\ 
 \textit{Improvement} & & \textit{(9.9\%)} & \textit{(113.1\%)} & \textit{(7.6\%)} & \textit{(40.8\%)} & \textit{(110.7\%)} \\ \addlinespace[0.5em] \hline \addlinespace[0.5em]
\multirow{2}{*}{\shortstack[l]{Spray Bottle\\Pink Button}} & GraspSplats~\cite{jigraspsplats} & \textbf{0.0028} & 0.4001 & 12.7509 & 0.2529 & 7.9788 \\
 & \textbf{Ours} & \textbf{0.0028} & \textbf{0.4134} & \textbf{12.5954} & \textbf{0.2534} & \textbf{8.0849} \\ 
 \textit{Improvement} & & \textit{(0.0\%)} & \textit{(3.3\%)} & \textit{(1.2\%)} & \textit{(0.2\%)} & \textit{(1.3\%)} \\ \bottomrule
\end{tabular}%
}
\end{table}

\textbf{Quantitative evaluation.}
We evaluate the proposed module through two dimensions: part-level semantic query performance and 3D functional keypoint localization accuracy.

First, we quantitatively evaluate the positioning accuracy of part-level semantic queries. As shown in Table \ref{tab:performance_comparison}, our proposed enhancements achieve significant improvements across all evaluation metrics compared to the baseline method, GraspSplats~\cite{jigraspsplats}. For queries targeting relatively large components such as the ``Hammer Handle'', the Precision Energy (P\_{En}) increased substantially from $0.2798$ to $0.5962$ (a $113.1\%$ improvement), and the Normalized Scanpath Saliency (NSS) grew by $110.7\%$. 
These results indicate that the part-level heatmaps generated by our method exhibit exceptionally high focus, effectively addressing the issues of feature diffusion and background interference prevalent in the original method. 
Simultaneously, for queries involving fine-grained, extremely small parts like the ``Pink Button of the Spray Bottle'', our method maintains superior MAE and KLD scores. The experimental data fully demonstrate that by introducing the HSE module, the localization precision and spatial consistency of fine-grained parts across various scales can be significantly enhanced within 3D Gaussian Splatting (3DGS) scenarios.

Following the evaluation protocol of GaussianGrasper~\cite{zheng2024gaussiangrasper}, we adopt \textit{Localization Success Rate} (LSR) as the core metric. A localization is considered successful only when the predicted three-point topological structure simultaneously satisfies the required grasp positions and the task-specific geometric functional constraints. 

\begin{figure*}[!t]
    \centering
\includegraphics[width=0.9\linewidth]{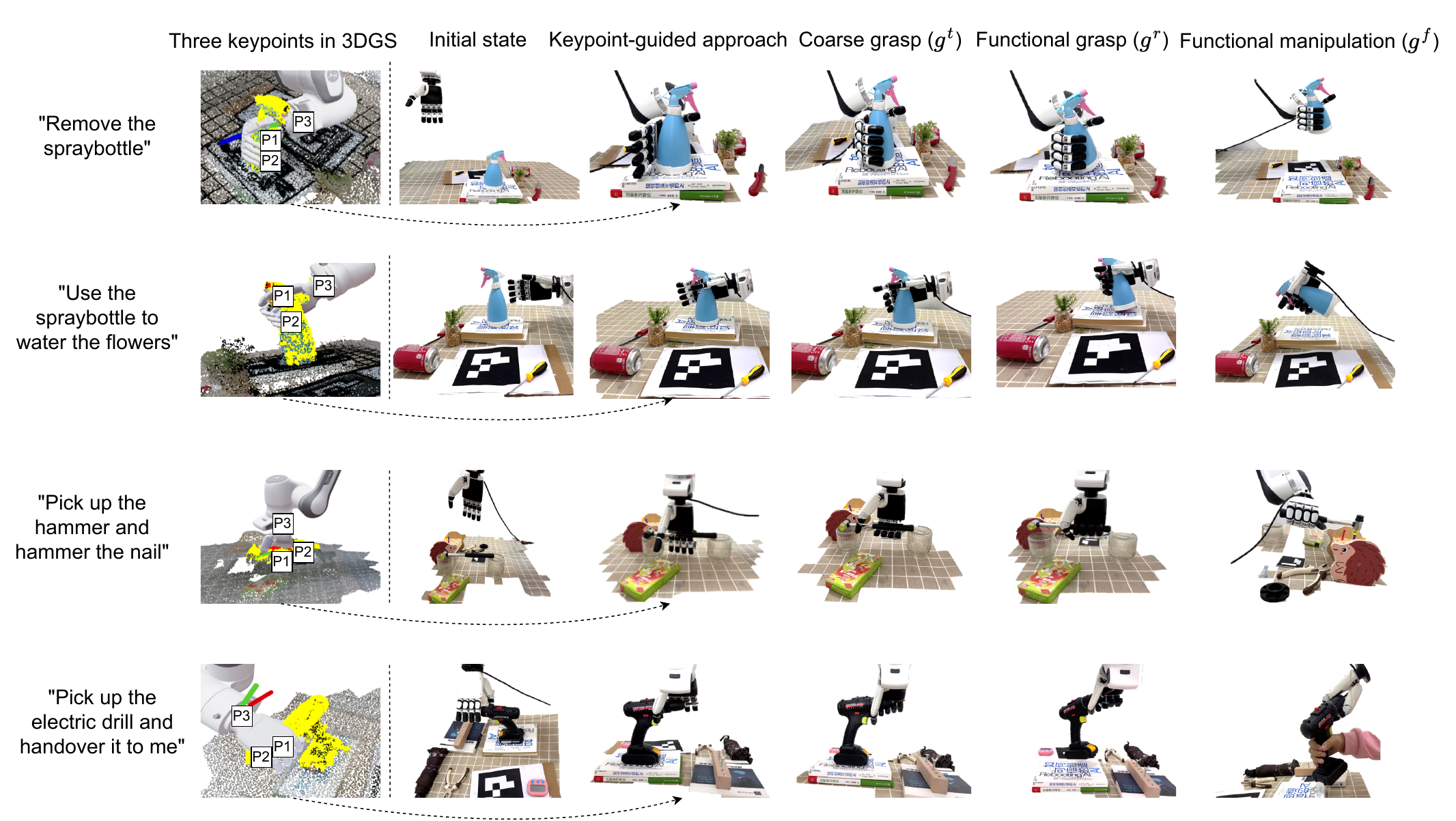}
    \caption{
    Real-world demonstrations of BLaFA on four language-guided dexterous manipulation tasks, illustrating 3D functional keypoint localization and subsequent grasping and manipulation.} 
    \label{fig:lva}
\end{figure*}

\begin{table}[t]
\centering
\caption{Comparison of localization success rates (LSR, \%) across four task categories.}
\label{tab:tri_lsr}
\renewcommand{\arraystretch}{1.2}
\begin{tabular}{l|cccc|c}
\toprule
\textbf{Method} & \textbf{Hold} & \textbf{Press} & \textbf{Open} &\textbf{Click} & \textbf{Mean LSR} \\ \midrule
MKA~\cite{yang2025multi}   & 50        & 20         & 10 & 10        & 22.5              \\ \hline
w/o Local Coordinate      & 50         & 0           & 0         & 50 & 25             \\ 
w/o HSE Module            & 62.5      & 50           & 50        & 50.0& 53.13              \\ \hline
\textbf{TriLocation (Ours)} & \textbf{75} & \textbf{50}  & \textbf{100} & \textbf{50}  & \textbf{68.75}    \\ \bottomrule
\end{tabular}
\end{table}

We evaluate our method across $6$ typical real-world scenes as shown in Fig.~\ref{fig:hardware}, covering $14$ representative task-tool combinations categorized into four functional intents: \textit{Hold} (including knife, spray bottle, drill, umbrella, hammer, pliers, bottle, and flashlight), \textit{Press} (drill and spray bottle), \textit{Open} (valve and bottle), and \textit{Click} (flashlight and mouse). 
\begin{figure}[t]
    \centering
\includegraphics[width=1\linewidth]{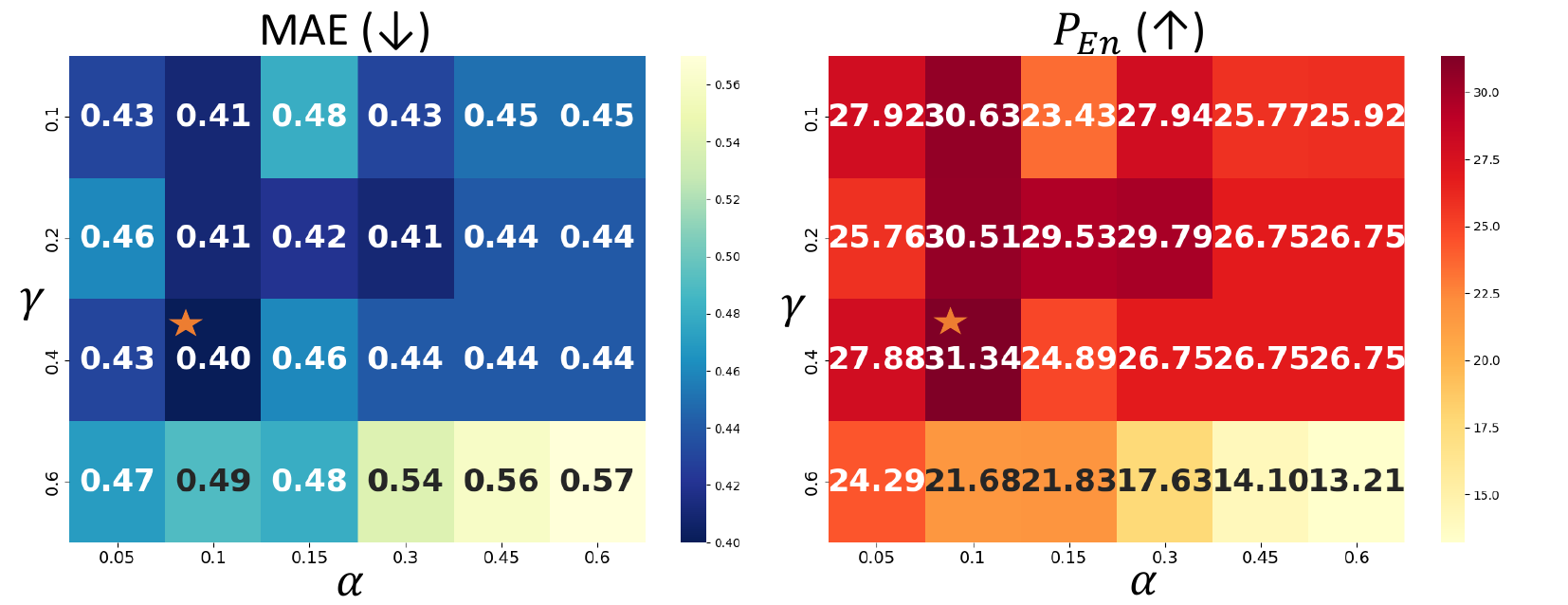}
    \caption{
    Hyperparameter analysis of $\alpha$ and $\gamma$. The optimal configuration is highlighted.
    }
    \label{fig:hypers}
\end{figure}

As shown in Table~\ref{tab:tri_lsr}, TriLocation achieves a mean LSR of $68.75\%$, calculated as the unweighted average across the four task categories. Unlike our method, MKA~\cite{yang2025multi} lacks open-vocabulary semantic query capabilities and relies on manual point-clicking to identify functional parts. MKA was evaluated over 10 trials per task-tool combination, whereas TriLocation was evaluated once per combination. Under these respective evaluation protocols, MKA achieves a mean LSR of $22.5\%$, with depth errors during 2D-to-3D lifting being a major challenge in complex environments.

Ablation studies further highlight the necessity of our proposed modules. Without the task-topology local coordinate frame constraint, the system fails to resolve the orientation for critical tasks like ``Press'' and ``Open'' (dropping to $0\%$ success rate in these categories), causing the mean LSR to drop sharply to $25\%$. 
Similarly, removing the Hierarchical Semantic Extraction (HSE) module leads to a decrease in LSR to $53.13\%$, as fine-grained part features become blurred by global object context. 
These results demonstrate that the integration of hierarchical mask constraints and task-topology reasoning effectively suppresses feature diffusion and ensures the high precision and spatial consistency of 3D fine-grained keypoint localization.

\textbf{Hyperparameter Analysis.} 
We evaluate the sensitivity of our module to the mask-to-box area ratio threshold $\alpha \in \{0.05, \dots, 0.6\}$ and padding ratio $\gamma \in \{0.1, \dots, 0.6\}$. 
As shown in Fig.~\ref{fig:hypers}, we record the $MAE$ and $P_{En}$ using the ``Flashlight Button'' as a representative fine-grained query. 

The results indicate that the optimal performance is achieved at $(\alpha=0.1, \gamma=0.4)$, yielding the lowest $MAE$ ($0.0040$) and highest $P_{En}$ ($0.3134$). 
We observe that a moderate padding ratio ($\gamma=0.4$) provides essential spatial context for the CLIP~\cite{radford2021learning} encoder, effectively suppressing semantic drift. 
However, excessively large values ($\gamma=0.6$) introduce background noise that contaminates the part features. Furthermore, the performance remains effective for $\alpha \leq 0.15$, but significantly degrades when $\alpha \geq 0.3$, suggesting that an overly restrictive area-ratio threshold might lead to the loss of critical part-level semantic information.

\subsection{BLaFA Performance in Real-world Environments}
Fig.~\ref{fig:lva} presents qualitative results of BLaFA on real-world language-guided manipulation tasks, ranging from simple object relocation to functional tool use, including ``Remove the spray bottle'', ``Use the spray bottle to water the flowers'', ``Pick up the hammer and hammer the nail'', and ``Pick up the electric drill and hand it over to me''.

As shown in Fig.~\ref{fig:lva}, the first column visualizes the three functional keypoints localized by TriLocation in the 3DGS scene. KGT3D+ uses these keypoints to construct the wrist pose $(R,T)$, guiding the robot from its initial state (second column) to the target grasping position (third column). The robot then performs an initial grasp according to the grasp type $g^t$ (fourth column), followed by functional finger interactions guided by the finger-role assignment $g^r$ (fifth column). Finally, the robot moves the grasped tool above the target object along a predefined motion, while the force label $g^f$ determines the force exerted by the functional fingers to accomplish the task (sixth column). For example, in the watering task, the index finger presses the spray bottle button to trigger spraying. These results demonstrate the ability of BLaFA to translate language-derived manipulation constraints into 3D grasp configurations and functional finger actions.

\begin{table}[!t]
    \centering
    \caption{Functional grasp success rate (FSR, \%) on two representative real-world task-tool combinations (10 trials each).}
    \label{tab:fsr_realworld}
    \begin{tabular}{lccc}
        \toprule
        Task-Tool & BLaFA & MKA~\cite{yang2025multi} & DP*~\cite{chi2025diffusion}\\
        \midrule
        Hold Spraybottle  & 80 & 40 & 50 \\
        Press Spraybottle & 50 & 10 & 10 \\
        \bottomrule
    \end{tabular}
\end{table}

\textbf{Comparative Analysis.}
We compare BLaFA with two baselines on two representative task-tool combinations: holding and pressing a spray bottle. MKA~\cite{yang2025multi} serves as an affordance-based grasping baseline, while DP* is adopted as a VLA-based policy baseline. DP* extends Diffusion Policy~\cite{chi2025diffusion} with an additional prediction head for 6-DoF dexterous hand joint control and is trained or fine-tuned using 20 teleoperation/data-glove demonstration trajectories per task category.

All methods are evaluated under the same sensing configuration and testing scenarios with identical task intents, while retaining their respective input and control mechanisms. Specifically, MKA relies on manual keypoint selection, whereas DP* uses task-specific demonstrations. Each task-tool combination is evaluated over 10 independent trials.

As shown in Table~\ref{tab:fsr_realworld}, BLaFA achieves the highest success rates in both tasks, reaching 80\% for holding and 50\% for pressing, compared with 40\% and 10\% for MKA, and 50\% and 10\% for DP*, respectively. The advantage is particularly evident in the pressing task, which requires precise alignment with functional components and finger-level interaction. These results highlight the benefits of explicitly modeling functional semantics and 3D contact geometry for real-world dexterous manipulation.

\section{Conclusions}

We proposed BLaFA, a structured framework for language-driven functional dexterous grasping. By introducing a semantic sextuple as an intermediate representation, BLaFA establishes explicit connections among natural-language task intentions, 3D functional contact structures, and dexterous grasp generation.
Specifically, KLP transforms natural-language instructions into structured manipulation constraints through knowledge-guided language parsing; TriLocation constructs hierarchical 3D semantic representations and localizes functional contact keypoints under task–tool topology constraints; and KGT3D+ maps the resulting geometric and semantic constraints to executable wrist poses and finger-level control commands. Experimental results demonstrate that BLaFA effectively translates natural-language instructions into task-aligned 3D functional contact structures and dexterous grasp configurations, providing a new approach for enabling robots to perform diverse tool-use tasks.

\bibliographystyle{IEEEtran}   
\bibliography{ref}

\end{document}